%% file: cais-2026-redpanda-adp.tex
\documentclass[sigconf,natbib=false]{acmart}
\AtBeginDocument{%
  }

\setcopyright{none}
\copyrightyear{2026}
\acmYear{2026}
\acmDOI{}
\acmConference[SAO '26]{Supporting Our AI Overlords, co-located with ACM CAIS '26}{May 26,
  2026}{San Jose, CA}
\acmISBN{}

\usepackage{tikz}

\RequirePackage[
  datamodel=acmdatamodel,
  style=acmnumeric,
  ]{biblatex}

\begin{document}

\title{The Importance of Out-of-Band Metadata for Safe \\ Autonomous Agents: The Redpanda Agentic Data Plane}

\author{Tyler Akidau, Tyler Rockwood, Johannes Brüderl, Marc Millstone}
\email{{takidau, rockwood, johannes, marc}@redpanda.com}
\affiliation{%
  \institution{Redpanda}
  \country{USA / Germany}}

\renewcommand{\shortauthors}{Akidau et al.}

\begin{abstract}
AI agents are increasingly expected to operate as digital employees:
accessing enterprise data, making decisions, and taking actions
autonomously. But agents are simultaneously \emph{less predictable} than
humans---prone to hallucination, misinterpretation, and
adversarial manipulation---and \emph{more technically capable}: with deep
system knowledge and high-throughput interfaces cascading damage at machine speed. This combination makes it unsafe to
rely on agents to faithfully interpret or propagate security-critical metadata
such as access policies, data classifications, and behavioral constraints.

We present the Redpanda Agentic Data Plane (ADP), an architecture built
around \emph{out-of-band metadata channels}: infrastructure
pathways that carry security context, policy signals, and audit
trails deterministically, entirely outside the agent's read and
write path and across heterogeneous infrastructure. These channels
enforce governance at every stage of
the agent lifecycle---scoping data access on the way in,
constraining actions during execution, and capturing
tamper-proof transcripts on the way out.

We demonstrate ADP with a multi-agent portfolio
rebalancing system in which autonomous agents monitor markets,
make trade decisions, and execute orders across isolated client
accounts---with per-client data scoping, trade approval
thresholds, and tamper-proof audit trails all enforced by
out-of-band channels the agents can neither see nor bypass.
\end{abstract}




\maketitle

\section{Introduction}

The vision for AI agents in the enterprise is ambitious: autonomous
software entities that access private data, reason over it, and take
consequential actions---functioning, in effect, as digital employees.
Like human employees, agents should have broad but appropriately
scoped access to organizational data. Enterprises already maintain
access control models for their human workforce; extending these to
agents is a natural starting point.

However, agents differ from humans in two important ways that
undermine this analogy. First, agents are \emph{less predictable}.
Large language models hallucinate, misinterpret instructions, and
are susceptible to prompt injection and jailbreaking. An agent may
inadvertently exceed its intended scope or be manipulated into doing
so. Second, agents are \emph{more technically capable}. Unlike a
human interacting through a keyboard, an agent has deep technical
knowledge and connects to production systems through
high-throughput, low-latency interfaces. A confused or compromised
agent can exfiltrate data, execute unauthorized transactions, or
cascade failures at machine speed.

This dual nature---less predictable yet more capable---means that
security-critical metadata cannot safely travel through channels the
agent can read, write, or interpret. Access policies embedded in
system prompts can be ignored. Data classifications passed as tool
parameters can be hallucinated away. Behavioral constraints encoded
in-band can be circumvented, whether
through adversarial manipulation or simple misinterpretation.

The architectural response we propose is \emph{out-of-band metadata
channels}: infrastructure-level pathways that propagate security
context, policy signals, and audit information deterministically,
entirely outside the agent's data path. These channels are not
advisory; they are enforced by the platform and invisible to the
agent. The agent cannot read a policy it should not know about,
cannot modify an access scope it did not set, and cannot tamper with
an auditing transcript it did not write.

Critically, enterprise data does not live in a single system. It
spans databases, APIs, message brokers, object stores, and SaaS
platforms. An effective out-of-band metadata plane must therefore
be interoperable by design, propagating security context across
heterogeneous infrastructure rather than requiring consolidation
into a single, managed, walled garden environment.

We present the Redpanda Agentic Data Plane (ADP), a system that embodies
these principles. ADP provides a unified runtime for agent-data
interaction in which out-of-band metadata channels operate at every
stage: scoping data access on the way in, constraining autonomous
actions during execution, and capturing tamper-proof audit trails on
the way out. We demonstrate ADP in a financial services scenario
where autonomous trading agents operate over per-user scoped
transactional data, with out-of-band guard\-rails that the agent can
neither see nor bypass.

\section{Problem: Why In-Band Metadata Fails}

Current approaches to agent governance typically embed policy
information in channels the agent itself processes: system prompts
that describe access restrictions, tool schemas that encode
permissions, or API responses that carry classification labels. This
approach inherits a fundamental assumption from conventionally
programmed systems---that the software consuming the metadata will
interpret and respect it deterministically.

For agents, this assumption fails on both sides of the
predictability-capability spectrum.

\paragraph{Unpredictable agents ignore metadata.}
An agent instructed via system prompt to ``only access records for
user~A'' may, through hallucination or prompt injection, query
records for user~B. If the enforcement mechanism is the agent's own
compliance with the instruction, there is no backstop. Similarly, a
data classification label returned as a field in a tool response
depends on the agent correctly parsing, interpreting, and acting on
that label---each step an opportunity for failure.

\paragraph{Capable agents circumvent metadata.}
A technically sophisticated agent with direct API access can craft
requests that bypass in-band restrictions. If a tool's access scope
is communicated through parameters the agent constructs, a
compromised agent can simply construct different parameters. The
higher the throughput of the agent's interface, the faster this
exploitation can occur and the larger its blast radius.

\paragraph{Audit trails must be tamper-proof.}
If transcript collection relies on the agent
reporting its own actions, a compromised agent can omit, fabricate,
or alter records. Out-of-band transcript capture---where the
infrastructure records interactions independently of the
agent---is the only reliable foundation for accountability. In
multi-agent pipelines, the problem compounds: each agent
boundary is a point at which in-band identity context can be
forged, dropped, or misrepresented by a compromised
intermediary---a classic instance of the confused deputy
problem~\cite{hardy1988confused}.

\medskip
These failure modes are not hypothetical. They are inherent to any
architecture where the enforcement boundary and the agent's
operational boundary overlap. Our threat model treats the agent
itself---including the LLM, the agent's runtime process, and any
external tools it invokes---as untrusted. An agent may be confused
(hallucinating, misinterpreting instructions) or actively compromised
(via prompt injection, poisoned tool output, or malicious upstream
context). In either mode, we assume it may attempt to read data
outside its authorized scope, construct requests that exceed its
granted authority, or suppress records of its own actions. The
platform infrastructure---gateways, message broker, identity
provider, and transcript store---is trusted to enforce policy as
configured; administrators configuring that policy are trusted, and
policies are assumed to be correctly specified. Out-of-band channels
defend against the agent, not against a compromised platform operator
or a policy author who grants excessive scope in the first place.

\section{Solution: Out-of-Band Metadata Channels}
\label{sec:solution}

We define an \emph{out-of-band metadata channel} as an
infrastructure-level pathway that carries security-critical
information alongside---but entirely outside of---the agent's data
path. Out-of-band channels have three defining properties:

\begin{enumerate}
\item \textbf{Agent-inaccessible.} The agent cannot read or write
  the channel. Policy, identity context, and audit signals propagate
  through infrastructure components (gateways, proxies, message
  brokers) that the agent interacts with but does not control.

\item \textbf{Deterministic.} Channel behavior is defined by
  configuration, not inference. A policy that limits an agent to
  10~trades per hour is enforced by a counter in the gateway, not
  by the agent's interpretation of a rate limit instruction.

\item \textbf{Interoperable.} Channels propagate across system
  boundaries \emph{and} agent boundaries. Security context
  established at an identity provider flows through an API gateway,
  across a message broker, through a multi-agent pipeline, into a
  database query layer, and back, without requiring all components
  to live within a single platform or agents to relay context in
  their payloads.
\end{enumerate}

These channels apply at every stage of the agent lifecycle:

\paragraph{Inbound: scoped data access.}
When an agent requests data, the metadata channel carries identity
and authorization context from the authentication layer to the data
source. The data source enforces row-level or resource-level
filtering based on this context. The agent receives only the data
it is permitted to see; it never observes the filtering logic or
the existence of data outside its scope.

\paragraph{Execution: constrained actions.}
When an agent takes an action---placing a trade, sending a message,
modifying a record---the metadata channel carries policy constraints
to the execution layer. Guard\-rails such as rate limits, value
thresholds, and approval requirements are enforced by the
infrastructure. The agent issues the action; the platform decides
whether it proceeds.

\paragraph{Outbound: tamper-proof audit.}
Every interaction between the agent and the data plane is recorded
by the infrastructure into an audit transcript. The agent does not
participate in this recording and cannot suppress or alter it. These
transcripts are themselves access-controlled: an agent tasked with
analyzing operational logs can only see transcripts it has been
explicitly granted access to.

\section{System: The Agentic Data Plane}

The Redpanda Agentic Data Plane (ADP) is a runtime and control plane for
agent-data interaction that implements out-of-band metadata channels
across four infrastructure layers.

\paragraph{Access control layer.}
An \emph{AI Gateway} handles LLM routing, token budgeting, and
failover, while an \emph{MCP Gateway} enforces tool-level policies
including PII redaction and resource filtering. Both gateways
integrate with enterprise identity providers to propagate
per-agent, per-user authorization context without exposing it to the
agent. Input and output guard\-rails---content filtering, prompt
injection detection, response validation---are applied at the
infrastructure boundary before the agent processes or emits data.

\paragraph{Data connectivity layer.}
ADP connects to heterogeneous data sources---managed and external
MCP servers, REST APIs, databases, object stores, and streaming
platforms---through adapters that propagate out-of-band metadata
across system boundaries. Agents interact with a uniform tool
interface; the metadata plane handles cross-system policy
enforcement transparently.

\paragraph{Agentic compute layer.}
Agents need persistent state and code execution. ADP provides
these via sandboxes where network isolation, resource limits, and
identity-scoped access are enforced by the infrastructure rather
than the agent. Gateway-governed MCP tools are projected into each
sandbox as executable actions, keeping compute and state within
ADP's governance envelope. The same out-of-band channels govern
ADP-managed and external agents alike.

\paragraph{Accountability layer.}
Every agent-data interaction is automatically recorded into
structured transcripts. These transcripts are collected by the
infrastructure, not reported by the agent, ensuring tamper-proof
provenance. Transcript access is itself governed by the same
out-of-band metadata channels, enabling agents to analyze
operational logs only within their authorized scope.

\section{Demo: Autonomous Wealth Management}

\input{figures/data-flow}

We demonstrate ADP with a live multi-agent wealth management
system\footnote{Accompanying demo video: https://tinyurl.com/redpanda-adp-demo-for-cais} that continuously monitors markets and rebalances client
portfolios.
The system manages multiple isolated client accounts
simultaneously, making autonomous trading decisions subject to
infrastructure-enforced guard\-rails that no agent can observe or
circumvent. The agents (LangChain on Google Cloud Run), client
data (MongoDB), and external services (Alpaca, Perplexity)
all reside outside the ADP, demonstrating out-of-band governance
across heterogeneous infrastructure rather than within a single
managed platform.

\subsection{Agent Pipeline and Data Flow}

Three agents and a human approval interface work together (Figure~\ref{fig:dataflow}),
communicating exclusively through async message channels
rather than direct inter-agent calls---ensuring every message
passes through infrastructure where out-of-band metadata can be
attached, inspected, and enforced.

The \emph{Signal Agent} runs a continuous research loop per
client, querying external research APIs for market developments
and publishing structured discoveries to per-client message
channels. The \emph{Decision Agent} consumes these discoveries
in a tool-calling loop, inspecting portfolio positions, buying
power, and price history before producing a trade recommendation
with an estimated amount and written rationale, which it
publishes to a single output channel. The infrastructure routes
the message based on two out-of-band signals the agent neither
sees nor controls: the agent's identity scope determines which
client the order belongs to, and the estimated value, compared against a
configured threshold, determines whether it proceeds to autonomous
execution or is held for human approval.

An \emph{Approval App} presents pending above-threshold orders
for human review. Approved orders are forwarded to the execution
channel; denied orders are discarded with an audit record. The
\emph{Execution Agent} consumes from the execution channels,
submits orders to a brokerage API, and records the result.

\subsection{Out-of-Band Channels in Practice}

Each stage of this pipeline exercises a distinct out-of-band
metadata channel as described in Section~\ref{sec:solution}.

\paragraph{Scoped data access via the MCP Gateway.}
All agents access external services---research APIs,
brokerage accounts, price feeds---exclusively through MCP
servers governed by the ADP's MCP Gateway. Each agent
authenticates with an infrastructure-issued credential; the
gateway resolves this credential to a client scope and enforces
it on every tool call. When the Decision Agent invokes
\texttt{get-positions}, it does not pass a client identifier as
a parameter for the LLM to construct---the gateway injects the
correct scope from the out-of-band identity context. An agent
cannot request another client's portfolio because the filtering
happens in the gateway, not in the agent's tool call. When the
gateway denies a request, the agent receives a standard tool-call
error; retry and escalation behavior is application-level, keeping
the out-of-band channel free of agent-driven negotiation.

\paragraph{Model-agnostic governance via the AI Gateway.}
Every LLM call routes through an AI Gateway that interposes
out-of-band controls between agent and model provider.
Agents authenticate with OIDC credentials rather than shared
API keys; the gateway resolves each identity to per-agent policies
and forwards the request to the appropriate model. Because the
agent never holds a provider key, it cannot circumvent the gateway.

This indirection enables governance mechanisms that operate
entirely out-of-band: \emph{budget controls} cap token consumption
per agent, preventing runaway loops from exhausting spend;
\emph{rate limits} contain the blast radius of a compromised agent;
\emph{input and output guard\-rails} detect prompt injection and
validate responses at the gateway boundary without the agent's
awareness; and \emph{dynamic routing} directs requests across
models and providers based on cost, latency, or policy---changing
backends without agent modification.

\paragraph{Scoped messaging via per-agent credentials.}
Each agent authenticates to the message infrastructure with its
own credential, and access control policies restrict which
channels each credential may produce to or consume from. The
Signal Agent can write to signal channels but cannot read from
or write to order channels. The Execution Agent can consume from
execution channels but cannot produce signals or approve orders.
These boundaries are enforced by the infrastructure; the agents
are unaware of them.

\paragraph{Constrained execution via the autonomy threshold.}
The dollar threshold that separates autonomous execution from
human-gated approval is an infrastructure policy, not an
instruction in the agent's prompt. The routing decision is made
by deterministic logic in the data plane after the agent has
produced its recommendation. The agent's output is the same
regardless of the threshold; the infrastructure decides where
that output goes. This separation ensures that a compromised or
confused Decision Agent cannot bypass approval by misreporting
an order's estimated value---the enforcement point is downstream
of the agent.

\paragraph{Tamper-proof audit via infrastructure-collected traces.}
Every agent interaction is captured as a distributed trace by the
infrastructure, not by agent self-reporting. The trace propagates
end-to-end: from the initial market research query, through every
LLM call routed through the AI Gateway, through each tool call
routed through the MCP Gateway, across the
message channels that connect agents, through the human approval
step (if any), to the final brokerage fill confirmation. Because
the trace context propagates out-of-band via standard headers
(W3C Trace Context~\cite{w3c2021tracecontext}) injected by the platform at each hop, the
agents cannot selectively omit steps or fabricate provenance.
The resulting trace constitutes a durable, tamper-proof record of
what each agent saw, decided, and did---the compliance artifact a
regulator would request.

\section{Related Work}

\paragraph{Agents cannot self-enforce safety.}
A growing body of work demonstrates that LLM agents routinely
violate safety constraints communicated through in-band channels.
ToolEmu~\cite{ruan2024toolemu} shows that agents commonly take
unsafe actions when given tool access, even in high-stakes scenarios.
AgentDojo~\cite{debenedetti2024agentdojo} demonstrates that prompt
injection can hijack agent tool calls in realistic scenarios.
Most directly, Cartagena and Teixeira~\cite{cartagena2026mindthegap}
show that alignment which suppresses harmful \emph{text} does not
suppress harmful \emph{tool calls}---safety trained into the
model's language behavior does not transfer to its actions.
These failure modes are well-documented~\cite{owasp2025llm}
and motivate removing enforcement
from the agent's reasoning path entirely.

\paragraph{Governance architectures and access control.}
Several recent works propose architectural responses to agent
safety failures.
Rajagopalan and Rao~\cite{rajagopalan2026authenticated} introduce
\emph{authenticated workflows} that enforce intent and integrity
at agent boundary crossings via cryptographic mechanisms---the
closest prior work to ours, though their security context
propagates in-band with each operation rather than through
agent-inaccessible infrastructure channels.
Shi et al.~\cite{shi2025trustauth} systematize the
\emph{trust-authorization mismatch}: static permissions are
structurally decoupled from an agent's fluctuating runtime
trustworthiness, precisely the gap out-of-band channels address.
Ji et al.~\cite{ji2026mac} apply mandatory access control to
prevent privilege escalation, enforcing policies outside the
agent's control.
Kim et al.~\cite{kim2025promptflow} enforce prompt flow integrity
to prevent cross-agent privilege escalation.
MiniScope~\cite{zhu2025miniscope} and
Progent~\cite{shi2025progent} enforce least-privilege policies
over tool calls through permission hierarchy reconstruction and
a programmable DSL, respectively.
Faramesh~\cite{faramesh2026} and the Generative Application
Firewall~\cite{vendrell2026gaf} introduce execution control
planes analogous to service mesh sidecars and web application
firewalls, respectively.
These works share our premise that enforcement must be external
to the agent; our contribution is framing out-of-band metadata
channels as the unifying architectural primitive.

\paragraph{MCP security, data flow, and audit.}
The Model Context Protocol~\cite{anthropic2024mcp} standardizes
tool exposure to agents but does not define governance or access
control within the protocol, though later revisions introduced
transport-level authentication.
South et al.~\cite{south2025delegation} extend OAuth~2.0 for scoped,
revocable agent delegation credentials, while
SMCP~\cite{hou2026smcp} and Jamshidi et
al.~\cite{jamshidi2025mcpsecurity} address MCP security at the
protocol level, including tool poisoning and adversarial attacks.
Summers et al.~\cite{summers2025vibe} argue that data flow control
should shift from agents to infrastructure, drawing an analogy to
how validation migrated from applications to the DBMS---a closely
related insight.
Garby et al.~\cite{garby2026llmbda} introduce a formal calculus
for information flow in agent conversations, proving
noninterference properties that injected prompts violate.
For accountability, AgentTrace~\cite{alsayyad2026agenttrace}
proposes continuous trace capture for agent
observability---a design whose rationale we share, since any
individual LLM interaction may be the one that matters for
compliance.
VET~\cite{grigor2025vet} and Doshi et
al.~\cite{doshi2026verifiable} provide cryptographically
verifiable execution traces and formal safety specifications
for tool-call sequences, respectively.
Our work differs in capturing complete audit trails at
the infrastructure layer---independent of both the agent and
the host---and in unifying access control, action constraints,
and audit under a single out-of-band metadata abstraction.

\paragraph{Architectural precedent.}
The closest precedent outside AI is the service
mesh~\cite{nist2024servicemesh} (e.g., Istio/Envoy), where
authentication, authorization, and telemetry are enforced by
sidecar proxies that applications cannot bypass. ADP applies
this pattern to agent-data interaction.
In industry, StrongDM's Leash
framework~\cite{strongdm2025leash, strongdm2025runtime}
independently validates the out-of-band principle, enforcing
agent policies at the OS kernel level via eBPF---complementing
the data-plane-level enforcement we describe here.

\section{Conclusion}

As AI agents assume the responsibilities of digital employees,
the infrastructure that mediates their access to enterprise data
must enforce governance guarantees that no agent---well-intentioned
or compromised---can undermine. Out-of-band metadata channels
provide this guarantee by moving security context, policy
enforcement, and audit capture entirely outside the agent's
operational boundary.

The Redpanda Agentic Data Plane demonstrates this architecture is
practical: out-of-band channels can scope data access, constrain
autonomous actions, and produce tamper-proof audit trails across
heterogeneous enterprise infrastructure, without sacrificing
broad data access that makes agents valuable in the first place.
Quantifying the latency and cost overhead of gateway-mediated
enforcement across production workloads is a natural direction for
future evaluation.


\printbibliography

\end{document}

%% file: figures/data-flow.tex

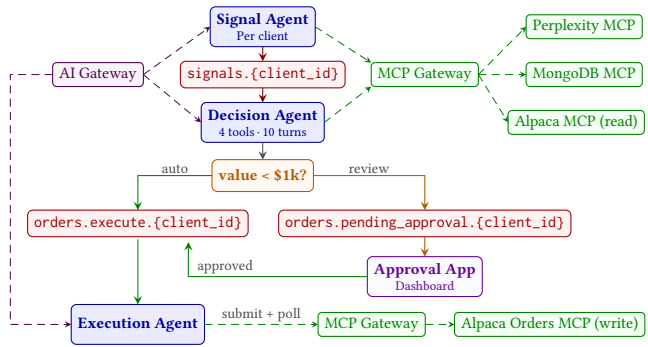
\begin{figure}[t]
\centering
\begin{tikzpicture}[
  x=1pt, y=-1pt,
  >=stealth,
  every node/.style={inner xsep=2.5pt, inner ysep=1pt, outer sep=0pt},
  agent/.style={draw=blue!60!black, fill=blue!8, rounded corners=2pt,
    minimum height=15pt, font=\fontsize{6}{7}\selectfont\bfseries,
    text=blue!60!black, align=center},
  mcp/.style={draw=green!50!black, fill=green!5, rounded corners=1.5pt,
    minimum height=9pt, font=\fontsize{6}{7}\selectfont,
    text=green!50!black},
  gateway/.style={draw=violet!60!black, fill=violet!5, rounded corners=1.5pt,
    minimum height=9pt, font=\fontsize{6}{7}\selectfont,
    text=violet!60!black},
  topic/.style={draw=red!60!black, fill=red!5, rounded corners=2pt,
    minimum height=10pt, font=\fontsize{6}{7}\selectfont\ttfamily,
    text=red!60!black},
  policy/.style={draw=orange!70!black, fill=orange!5, rounded corners=2pt,
    minimum height=12pt, font=\fontsize{6}{7}\selectfont\bfseries,
    text=orange!70!black},
  human/.style={draw=red!40!blue, fill=red!3!blue!2, rounded corners=2pt,
    minimum height=15pt, font=\fontsize{6}{7}\selectfont\bfseries,
    text=red!40!blue, align=center},
  datastore/.style={draw=gray!70, fill=gray!8, rounded corners=2pt,
    minimum height=11pt, font=\fontsize{6}{7}\selectfont,
    text=gray!60!black},
  dataflow/.style={->, thin, gray!60!black},
  toolcall/.style={->, densely dashed, ultra thin},
  topicflow/.style={->, thin, red!50!black},
  lbl/.style={font=\fontsize{5.5}{6}\selectfont, text=gray!60!black},
]


\node[gateway] (gw) at (15, 30) {AI Gateway};


\node[agent] (signal) at (77, 12)
  {Signal Agent\\[-1pt]{\fontsize{5}{6}\selectfont\mdseries Per client}};

\draw[toolcall, violet!60!black] (gw.east) -- (signal.west);


\node[mcp] (mcpgw) at (138, 30) {MCP Gateway};


\node[mcp] (perp) at (198, 12) {Perplexity MCP};
\node[mcp] (mongo) at (198, 30) {MongoDB MCP};
\node[mcp] (alpread) at (195, 48) {Alpaca MCP (read)};

\draw[toolcall, green!50!black] (signal.east) -- (mcpgw.north west);
\draw[toolcall, green!50!black] (mcpgw.east) -- (perp.west);
\draw[toolcall, green!50!black] (mcpgw.east) -- (mongo.west);
\draw[toolcall, green!50!black] (mcpgw.east) -- (alpread.west);


\node[topic] (sigtopic) at (77, 30)
  {signals.\{client\_id\}};
\draw[topicflow] (signal.south) -- (sigtopic.north);


\node[agent] (decision) at (77, 48)
  {Decision Agent\\[-1pt]{\fontsize{5}{6}\selectfont\mdseries 4 tools\,·\,10 turns}};
\draw[topicflow] (sigtopic.south) -- (decision.north);

\draw[toolcall, violet!60!black] (gw.east) -- (decision.west);
\draw[toolcall, green!50!black] (decision.east) -- (mcpgw.south west);


\node[policy] (threshold) at (77, 68)
  {value < \$1k?};
\draw[dataflow] (decision.south) -- (threshold.north);


\node[topic] (exectopic) at (30, 86)
  {orders.execute.\{client\_id\}};
\node[topic] (pendtopic) at (138, 86)
  {orders.pending\_approval.\{client\_id\}};

\draw[topicflow, green!50!black] (threshold.west) -| (exectopic.north)
  node[lbl, pos=0.25, above] {auto};
\draw[topicflow, orange!70!black] (threshold.east) -| (pendtopic.north)
  node[lbl, pos=0.25, above] {review};


\node[human] (approval) at (138, 106)
  {Approval App\\[-1pt]{\fontsize{5}{6}\selectfont\mdseries Dashboard}};
\draw[topicflow, orange!70!black] (pendtopic.south) -- (approval.north);

\draw[topicflow, green!50!black] (approval.west) -- (49, 106) -- (49, 93)
  node[lbl, pos=0.3, right] {\,approved};


\node[agent] (exec) at (30, 124)
  {Execution Agent};
\draw[topicflow, green!50!black] (30, 92) -- (exec.north);

\draw[toolcall, violet!60!black] (gw.west) -- (-18, 30) -- (-18, 124) -- (exec.west);

\node[mcp] (mcpgw2) at (118, 124) {MCP Gateway};
\node[mcp] (alpwrite) at (185, 124)
  {Alpaca Orders MCP (write)};
\draw[toolcall, green!50!black] (exec.east) -- (mcpgw2.west)
node[lbl, midway, above] {submit + poll};
\draw[toolcall, green!50!black] (mcpgw2.east) -- (alpwrite.west);

\end{tikzpicture}
\caption{Autonomous wealth management demo. Agents (blue) communicate
  exclusively through asynchronous messaging channels (red), never directly;
  out-of-band metadata propagates transparently. MCP
  tools (green) route through the MCP Gateway and LLM calls (purple) through
  the AI Gateway, both enforcing out-of-band policies. The autonomy threshold
  (orange) is an infrastructure routing decision that agents cannot observe
  or affect.}
\label{fig:dataflow}
\end{figure}